\def\year{2021}\relax
\documentclass[letterpaper]{article} 
\usepackage{aaai21}  
\usepackage{times}  
\usepackage{helvet} 
\usepackage{courier}  
\usepackage[hyphens]{url}  
\usepackage{graphicx} 
\usepackage{natbib}  
\usepackage{caption} 
\usepackage{cancel}
\usepackage{makecell}
\usepackage[dvipsnames]{xcolor}
\usepackage{hhline}
\usepackage[switch]{lineno} 

\usepackage{multirow}

\newcommand{\cut}[1]{}

\newcommand{\citenoun}[1]{{\citeauthor{#1} \shortcite{#1}}}

\title{Robustness to Spurious Correlations in Text Classification via Automatically Generated Counterfactuals}
\author {
        Zhao Wang\textsuperscript{\rm 1} \textnormal{and}
        Aron Culotta \textsuperscript{\rm 2}\\
}

\affiliations {
    \textsuperscript{\rm 1} Department of Computer Science, Illinois Institute of Technology, Chicago, IL \\
    \textsuperscript{\rm 2} Department of Computer Science, Tulane University, New Orleans, LA \\
    zwang185@hawk.iit.edu, aculotta@tulane.edu
}

\begin{document}


\maketitle
\begin{abstract}
\begin{quote}
Spurious correlations threaten the validity of statistical classifiers. While model accuracy may appear high when the test data is from the same distribution as the training data, it can quickly degrade when the test distribution changes. For example, it has been shown that classifiers perform poorly when humans make minor modifications to change the label of an example. One solution to increase model reliability and generalizability is to identify causal associations between features and classes. In this paper, we propose to train a robust text classifier by augmenting the training data with automatically generated counterfactual data. We first identify likely causal features using a statistical matching approach. Next, we generate counterfactual samples for the original training data by substituting causal features with their antonyms and then assigning opposite labels to the counterfactual samples. Finally, we combine the original data and counterfactual data to train a robust classifier. Experiments on two classification tasks show that a traditional classifier trained on the original data does very poorly on human-generated counterfactual samples (e.g., 10\%-37\% drop in accuracy). However, the classifier trained on the combined data is more robust and performs well on both the original test data and the counterfactual test data (e.g., 12\%-25\% increase in accuracy compared with the traditional classifier). Detailed analysis shows that the robust classifier makes meaningful and trustworthy predictions by emphasizing causal features and de-emphasizing non-causal features.

\end{quote}
\end{abstract}

\noindent 

\section{Introduction}
Despite the remarkable performance machine learning models have achieved in various tasks, studies have shown that statistical models typically learn correlational associations between features and classes, and model validity and reliability are threatened by spurious correlations. Examples include: a sentiment classifier learns that ``Spielberg'' is correlated with positive movie reviews~\cite{wang-culotta-2020-identifying}; a toxicity classifier learns that ``gay'' is correlated with toxic comments~\cite{Ellery2017toxic}; a medical system learns that the disease is associated with patient ID~\cite{Shachar2011Leakage}; an object detection system recognizes a sheep based on the grass in the background~\cite{ghorbani2019interpretation}. If these kinds of spurious correlations are built into the model during training time, the model could fail when test data has a different distribution or even on samples with minor changes, and the predictions will be problematic and suffer from algorithm fairness or trust issues.  


One solution to achieve robustness is to learn causal associations between features and classes.  E.g., in the sentence \textit{``This was a free book that sounded boring to me''}, the word most responsible for the label being negative is ``boring'' instead of ``free''. Identifying causal associations provides a way to build more robust and generalizable models.


Recent works try to achieve robustness with the aid of human-in-the-loop systems. \citenoun{srivastava2020robustness} present a framework to make models robust to spurious correlations by leveraging human common sense of causality. They augment training data with crowd-sourced annotations about reasoning of possible shifts in unmeasured variables and finally conduct robust optimization to control worst-case loss. Similarly, \citenoun{kaushik2019learning} ask humans to revise documents with minimal edits to change the class label, then augment the original training data with the counterfactual samples. Results show that the robust classifier is less sensitive to spurious correlations. While these prior works show the potential of using human annotations to improve model robustness, collecting such annotations can be costly.

In this paper, we propose to train a robust classifier with automatically generated counterfactual samples. Specifically, we first identify likely causal features using the closest opposite matching approach and then generate counterfactual training samples by substituting causal features with their antonyms and assigning opposite labels to the newly generated samples. Finally, we combine the original training data with counterfactual data to train a more robust classifier.

We experiment with sentiment classification tasks on two datasets (IMDB movie reviews and Amazon kindle reviews). For each dataset, we have the original training data and testing data, and additional human-generated counterfactual testing data. We first train a traditional classifier using the original data, which performs poorly on the counterfactual testing data (i.e., 10\%-37\% drop in accuracy). Then, we train a robust classifier with the combination of original training data and automatically-generated counterfactual training data, and it performs well on both the original testing data and the counterfactual testing data (i.e., 12\% - 25\% absolute improvement over the baseline). Additionally, we consider limited human supervision in the form of human-provided causal features, which we then use to generate counterfactual training samples. We find that a small number of causal features (e.g., 50) results in accuracy that is comparable to a model trained with $1.7K$ human-generated counterfactual training samples from previous work.


\section{Related Work}
Spurious correlations are problematic and could be introduced in many ways. \citenoun{sagawa2020investigation} investigate how overparameterization exacerbates spurious correlations. They compare overparameterized models with underparameterized models and show that overparameterization encodes spurious correlations that do not hold in worst-group data. \citenoun{kiritchenko2018examining} showed that training data imbalances can lead to unintended bias and unfair applications (e.g., bias towards gender, race). Besides that, data leakage \cite{Melissa2011causal} and distribution shift between training data and testing data \cite{Joaquin2009datasetshift} are particularly challenging and hard to detect as they introduce spurious correlations during model training and hurt model performance when deployed. Another new type of threat is backdoor attack \cite{Dai2019ABA}, where an attacker intentionally poisons a model by injecting spurious correlations into training data and manipulating model performance by specific triggers.

A growing line of research explores the challenges and benefits of using causal inference to improve model robustness. \citenoun{wooddoughty2018challenges} uses text classifiers in causal analyses to address issues of missing data and measurement error. \citenoun{keith2020text} introduce methods to remove confounding from causal estimates. \citenoun{paul-2017-feature} proposes a propensity score matching method to learn meaningful causal associations between variables. \citenoun{jia2019certified} consider label preserving transformations to improve model robustness to adversarial perturbations with Interval Bound Propagation. ~\citenoun{Virgile2018confounding} address the issue of spurious correlations by doing back-door adjustment to control for known confounders. \citenoun{wang-culotta-2020-identifying} train a classifier to distinguish between spurious features and genuine features, and gradually remove spurious features to improve worst-case accuracy of minority groups.

Recent works investigate how additional human supervision can reduce spurious correlations and improve model robustness. \citenoun{Melissa2011causal} and \citenoun{sap2018atomic} show that humans achieve high performance on commonsense causal reasoning and counterfactual tasks. \citenoun{Zaidan2008ModelingAA} ask annotators to provide rationales as hints to guide classifiers paying attention to relevant features. \citenoun{lu2018gender} and \citenoun{zmigrod-etal-2019-counterfactual} use counterfactual data augmentation to mitigate bias. \citenoun{ribeiro2020accuracy} evaluate model robustness using generated counterfactuals that requires significant human intervention (either by specifying substitution terms or generating templates and labeling examples). \citenoun{garg2019counterfactual} presume a predefined set of 50 counterfactually fair tokens and augment the training data with counterfactuals to improve toxicity classifier fairness.


While recent works have proposed the idea of generating and augmenting with counterfactuals for robust classifications, the main contributions of this paper are as follows:
\begin{itemize}
    \item We propose to discover likely causal features using statistical matching techniques.
    \item Using these features, we automatically generate counterfactual samples by substituting causal features with antonyms, which significantly reduces human effort.
    \item We conduct experiments demonstrating the improved robustness of the resulting classifier to spurious features.
    \item We conduct additional analyses to show how the robust classifier increases the importance of causal features and decreases the importance of  spurious features. 
\end{itemize}

%


\section{Problem and Motivation}
To train a classification model, we fit a function $f(\cdot)$ with a set of labeled data and learn a map between input features and output labels. We consider a binary text classification task with the simple approach of logistic regression model\footnote{Our approach is model agnostic. We focus on logistic regression for interpretability and clarity.}: $f(x; \theta) = \frac{1}{1+e^{-\langle x, \theta \rangle}}$ using bag-of-words features. Specifically, each document is a sequence of words $d=\langle w_1 \ldots w_k \rangle$ that is transformed into a feature vector $x$ via one-hot representation $x=\langle x_1 \ldots x_V \rangle$ ($V$ is the vocabulary size), and has a binary label $y \in \{-1, 1\}$. The model is fit on a set of labeled documents $\mathcal{D}=\{(d_1,y_1) \ldots (d_n,y_n)\}$, and parameters are estimated by minimizing the loss function $\mathcal{L}$:  $\theta^* \leftarrow \arg\min_{\theta} \mathcal{L}(\cal{D}, \theta)$. We can examine the (partial) correlations between features and labels by model coefficients.

Spurious correlations are very common in statistical models and they could mislead classifiers. For example, in our experimental dataset of Amazon kindle reviews, the classifier learns that ``free'' has a strong correlation with negative sentiment because ``free'' has a high frequency in negative book reviews (e.g., ``This was a \textit{free} book that sounded boring to me''), and thus the classifier makes errors when predicting positive documents that contain ``free''.

Previous works have tried various methods to reduce spurious correlations (e.g., regularization, feature selection, back-door adjustment~\cite{Hoerl1970Regression, Forman2003, Virgile2018confounding}). However, a more direct solution is to learn meaningful causal associations between features and classes. 
While expressing causality in the context of text classification can be challenging, we follow the previous work~\cite{paul-2017-feature} to operationalize the definition of a causal feature as follows: term $w$ is a {\bf causal feature} in document $d$ if, all else being equal, one would expect $w$ to be a determining factor in assigning a label to $d$. For example, in the sentence ``This was a free book that sounded \textit{boring} to me'', ``boring'' is primarily responsible for the negative sentiment. In contrast, the term ``free'' itself does not convey negative sentiment. We consider ``boring'' as a causal term and ``free'' as a non-causal term ({\it term} refers to {\it word feature}).


Our approach in this paper is to first identify such causal features and then use them to automatically generate counterfactual training samples. Specifically, for a sample $(d, y)$, we get the corresponding {\bf counterfactual} sample $(d',y')$ by (i) substituting causal terms in $d$ with their antonyms to get $d'$, and (ii) assigning an opposite label $y'$ to $d'$. Let's consider the previous example to see how augmenting with counterfactual samples might work. Traditional classifiers trained on original data learns that ``free'' is correlated with the negative class due to its high frequency in negative book reviews. For every negative document containing ``free'', we generate one corresponding counterfactual document. The counterfactual sample for ``This was a free book that sounded \textit{boring} to me''(neg) would be ``This was a free book that sounded \textit{interesting} to me''(pos). When augmenting the original training data with counterfactual data, ``free'' would get equal frequency in both classes for the ideal case (i.e, if we could generate counterfactual samples for all documents containing ``free''). Thus, a classifier fit on the combined dataset should have a reduced coefficient for ``free'' and increased coefficients for ``boring'' and ``interesting.''

\section{Methods}

Our approach is a two-stage process: we first identify likely causal features and then generate counterfactual training data using causal features. To identify causal features, we consider the counterfactual framework of causal inference~\cite{winship1999estimation}. If word $w$ in document $d$ were replaced with some other word $w'$, how likely is it that the label $y$ would change? Since conducting randomized control trials to answer this question is infeasible, we instead use matching methods~\cite{imbens2004nonparametric,king2019propensity}. The intuition is as follows: if $w$ is a reliable piece of evidence to determine the label of $d$, we should be able to find a very similar document $d'$ that (i) does not contain $w$, and (ii) has the opposite label of $d$. For example, $(d,y)$ = (``This was a free book that sounded \textit{boring} to me'', neg) and $(d',y')$ = (``This was a free book that sounded \textit{interesting} to me'', pos) would be an ideal match where substituting causal term ``boring'' with another term ``interesting'' flips the label. While this is not a {\it necessary} condition of a causal feature (there may not be a good match in a limited training set), in the experiments below we find this to be a fairly precise approach to generate a small number of high-quality causal features.


The full steps of our approach are as follows:
\begin{enumerate}
    \item We first train an initial classifier and extract strongly correlated terms $\langle t_1 \ldots t_k \rangle$ as candidate causal features. E.g., for logistic regression model, we would extract features with high magnitude coefficients. For more complex models, other transparency algorithms may be used~\cite{martens2014explaining}. 
    \item For each top term $t$ and a set of documents containing $t$: $D_t=\langle d_1 \ldots d_n \rangle$, we search for a set of matched documents $D'_t=\langle d_1' \ldots d_n' \rangle$ and get ${D_{match}}= \{(d_1, d_1', score_1) \ldots (d_n, d_n', score_n')\}$, where the score for each match is the context similarity of $d_i$ and $d_i'$. The matched documents have \textit{opposite} labels.
    \item Then for each term $t$ and its corresponding matching set $D_{match}$, we pick the tuple $(d_i, d_i', score_i)$ that has the highest similarity score as the \textit{closest opposite match}. We then identify likely causal features by picking those whose closest opposite matches have scores greater than a threshold (0.95 is used below).
    \item We use PyDictionary\footnote{https://github.com/geekpradd/PyDictionary} to get antonyms for causal terms.
    \item For each training sample, we generate its counterfactual sample by substituting causal terms with antonyms and assigning an opposite label to the counterfactual sample.
    \item Finally, we train a robust classifier using the combination of original training data and counterfactual data.
    
We provide more details on these steps below.

\end{enumerate}

\subsection{Identifying Likely Causal Features}

\begin{table*}[t]
    \small
    \centering
    \begin{tabular}{ | l| l | c |} 
     \hline
     \textbf{Original sentence} & \textbf{Matched sentence} & \textbf{Context similarity} \\
     \hline
    This was an \underline{\textit{amazing}} book. & This was a \underline{\textit{boring}} book & 0.977 \\
    
    It was a \underline{\textit{boring}} read. & The book was \underline{\textit{great}} and long. & 0.998 \\
    
    This short story was a \underline{\textit{disappointment}}. & This was a \underline{\textit{great}} short story. & 0.992 \\
    
    This is one of the \underline{\textit{funniest}} movies I have seen. & This is one of the \underline{\textit{worst}} movies I have ever seen. & 0.980 \\
    
    \underline{\textit{Fantastic}} film. & \underline{\textit{Terrible}} film. & 1.00 \\
    \hline
    \end{tabular}
    \caption{Examples of Closest Opposite Matches with Corresponding Context Similarity Scores}
    \label{tab.match}
\end{table*}

We expect causal features to have at least some correlations with the target class, so we first fit an initial binary classifier $f(x; \theta)$ on original training data $\mathcal{D}=\{(d_1,y_1) \ldots (d_n,y_n)\}$ and extract top terms $\langle t_1 \ldots t_k \rangle$ that have relatively large magnitude coefficients (e.g., $> 1$ in experiments below).

For a top term $t$ and a document $d$ containing $t$, we let $d[{\hat t}]$ represent the context of removing $t$ from $d$. We search for another document $d'$ that (i) has $t' \in d'$ and $t \notin d'$, where $t'$ is another top term, and (ii) $d'$ has the opposite label with $d$. We use a best match approach to search for $d'[{\hat t'}]$ that has highest semantic similarity to $d[{\hat t}]$ among all possible $d^*[{\hat t^*}]$:  $d' \leftarrow \arg\max_{d^*} \hbox{sim}(d[{\hat t}], d^*[{\hat t^*}])$. For a term $t$, we get a set of corresponding matches as ${D_{match}}= \{(d_1, d_1', score_1) \ldots (d_n, d_n', score_n)\}$, where the score for each match is the semantic similarity between $d[{\hat t}]$ and $d'[{\hat t'}]$. Each context is represented by concatenating the last four layers of a pre-trained BERT model (i.e., recommended by ~\cite{Devlin2019BERTPO}). We then select the match $(d_i, d_i', score_i)$ that has the highest score in $D_{match}$ as the {\bf closest opposite match} for $t$. Table~\ref{tab.match} shows examples of the closest opposite matches.



From the previous step, we get the closest opposite match for each top term. We then identify terms with closest opposite match scores greater than 0.95 as likely causal terms.

\begin{figure}[t]
	\centering
	\includegraphics[width=3.3in]{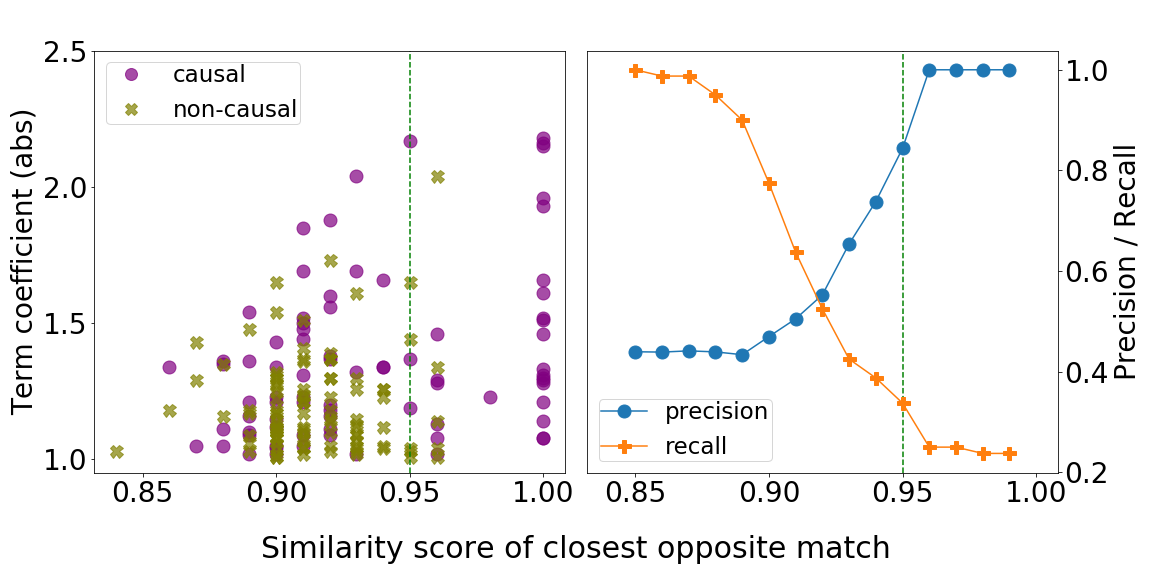}
	\caption{The ``closest opposite match'' score provides a high-precision indicator of causal features (IMDB dataset).}
	\label{fig:imdb_causal_terms}
\end{figure}

To evaluate the quality of this approach, the left panel of Figure~\ref{fig:imdb_causal_terms} shows terms annotated by a human as likely to be causal or not, plotted by both their closest opposite match scores as well as the magnitude of coefficients from the classifier trained on original data. We can see that terms with very high closest opposite match scores are very likely to be causal. Note that this is not necessarily the case for terms with high coefficients (y-axis). The high precision and low recall pattern is further supported by the right panel.

\subsection{Selecting Antonyms for Causal Terms}
After identifying causal terms, we search for their antonyms using PyDictionary. This package provides simple interfaces for getting meanings from WordNet and synonyms and antonyms from synonym.com. To reduce the noise of the returned antonyms, we require the antonyms to have opposite coefficients with the causal terms. Specifically, for each causal term $t$, we search for its antonyms by:
\begin{itemize}
    \item First, check the direct antonyms for $t$ and save those that satisfy the coefficient requirement as candidate antonyms.
    \item If no satisfying antonym is found, we then get synonyms for $t$ and iteratively search for each synonym's antonyms, and save the satisfied antonyms as candidate antonyms.
\end{itemize}
After these two steps, we get at least one candidate antonym for each causal term $t: \{a_1 \ldots a_k\}, k\geq1$. Table~\ref{tab.antonym} shows examples of the antonyms we get for causal terms.

\begin{table}
    \small
    \centering
    \begin{tabular}{ | l| l |} 
     \hline
     \textbf{Causal Term} & \textbf{Antonyms}  \\
     \hline
    fantastic: 1.638 & {unimpressive: -0.462; inferior: -0.644} \\ 
    awesome: 1.202	& {unimpressive: -0.462} \\
    pleasant: 1.106 & {unpleasant: -0.333} \\
    dull: -1.881 & {lively: 0.302; colorful: 0.252} \\
    boring:	-2.592	& {interesting: 0.734} \\
    \hline
    \end{tabular}
    \caption{Discovered antonyms for causal terms and corresponding coefficients from the initial classifier.}
    \label{tab.antonym}
\end{table}




\subsection{Generating Counterfactual Samples}
 
Next, for each training document $d$, we first identify all the causal terms in $d$: $\langle t_1 \ldots t_m \rangle$, and then substitute all causal terms with their corresponding antonyms. If a causal term has multiple candidate antonyms, we randomly pick one to substitute. We only generate counterfactuals for documents contain at least one causal term. Finally, we assign opposite labels to the generated samples. Table~\ref{tab.spurious_coef} shows examples of generated counterfactual sentences. While most substitutions result in reasonable sentences, future work may investigate more sophisticated language models to ensure fluency of the generated counterfactuals.


\subsection{Training a Robust Text Classifier}
We augment the original training data with the automatically generated counterfactual data to train a robust classifier. We perform experiments below to investigate how do causal terms affect the quantity and quality of automatically generated counterfactual samples.


\section{Data}\label{sec.data}


We perform sentiment classification experiments on the following two datasets.\footnote{Code and data available at: https://github.com/tapilab/aaai-2021-counterfactuals} Each dataset has human-generated counterfactual testing samples to provide benchmark performance for classifier robustness.

\textbf{IMDB movie reviews:} This dataset is sampled from the original IMDB dataset~\cite{pang2005seeing} and the conterfactual part is collected and published by~\citenoun{kaushik2019learning}. They randomly sampled $2.5K$ reviews with balanced class distributions and partition them into 1707 training, 245 validation, and 488 testing samples. Then they instruct Amazon Mechanical Turk workers to revise each document with minimum changes towards a counterfactual label, and finally collected $2.5K$ counterfactually-manipulated samples.

Each document of this dataset is a long paragraph. We are both interested in exploring classifier performance for long texts and short texts. So, we additionally create a version of this dataset segmented into single sentences. To do so, we first fit a binary classifier on the original data and identify strongly correlated terms as keywords. Then we split each original document into single sentences and keep those containing at least one keyword. Sentence labels are inherited from the original document labels. To justify the validity of this approach, we randomly sampled 500 sentences and manually checked their labels. The inherited labels were correct for 484 sentences (i.e., 96.8\% accuracy). We differentiate the IMDB dataset with long texts as \textbf{IMDB-L} and short texts as \textbf{IMDB-S}.


\textbf{Amazon Kindle reviews (Kindle):}  This dataset contains book reviews from the Amazon Kindle Store and each review has a rating ranges from 1-5~\cite{He_2016}. We label reviews with ratings \{4,5\} as positive and reviews with ratings \{1,2\} as negative, and then process this dataset to be single sentences following the approach used in IMDB. 

\textbf{Human edited counterfactuals:} For the IMDB dataset, we have the human-generated counterfactual training data and counterfactual testing data. For kindle dataset, we randomly select 500 samples as test data (comparable size with the test data from IMDB-L) and manually edit them to be counterfactual samples with the minimum edits.

\textbf{Ground truth causal terms:} We manually annotated a set of ground truth causal terms for each dataset. Specifically, we asked two student annotators to label a term as causal if, all else being equal, this term is a determining factor in assigning a label to a document. While there is some subjectivity in the annotation, we did a round of training to resolve disagreements prior to annotation and the final agreement was generally high for this task (e.g., 96\% raw agreement by fraction of labels that agree).

Table~\ref{tab.dataset} shows the basic data statistics. For the top terms, we select them by thresholding on the magnitude of coefficients. For IMDB-L, we use threshold 0.4, and for IMDB-S and Kindle, we use threshold 1.0.

\begin{table}[t]
    \small
    \centering
    \begin{tabular}{ |r|p{0.5cm}|p{.5cm}|p{.5cm}|p{.5cm}|p{.5cm}|p{.5cm}|} 
     \hline
     
     & \multicolumn{2}{c|}{\textbf{IMDB-L}} & \multicolumn{2}{c|}{\textbf{IMDB-S}} & \multicolumn{2}{c|}{\textbf{Kindle}}\\
     \hline
    &  {\it pos} & {\it neg} & {\it pos} & {\it neg} & {\it pos} & {\it neg} \\
    
    \hline
    Train & 856 & 851 & 4114 & 4059 & 5000 &5000\\
    \hline
    Test & 245 & 243 & 1144 & 1101 & 250 &250\\
    \hline
    
    Top terms & \multicolumn{2}{c|}{231} & \multicolumn{2}{c|}{198} & \multicolumn{2}{c|}{194}\\
    \hline
    Causal terms & \multicolumn{2}{c|}{282} & \multicolumn{2}{c|}{285} & \multicolumn{2}{c|}{264}\\
    \hline
    \end{tabular}
    \caption{Dataset summary }
    \label{tab.dataset}
\end{table}





\section{Experiments and Discussion}

\begin{table*}
    \centering
    \begin{tabular}{|p{2.8cm}|p{5.2cm}||p{.7cm}|p{.7cm}||p{.7cm}|p{.7cm}||p{.7cm}|p{.7cm}|} 
    \hline
     \multicolumn{2}{|c||}{\textit{Training data:}} & 
         \multicolumn{6}{c|}{\textit{Testing data}} \\
         \cline{3-8}
         \multicolumn{2}{|c||}{\textit{Original train samples + Counterfactual train samples}} &
         \multicolumn{2}{c||}{\textbf{IMDB-L}} & \multicolumn{2}{c||}{\textbf{IMDB-S}} & \multicolumn{2}{c|}{\textbf{Kindle}}\\
      
         \hline
         {\bf Counterfactual training samples} & {\bf Causal terms} & {\bf Orig} & {\bf CTF} & {\bf Orig} & {\bf CTF} & {\bf Orig} & {\bf CTF} \\
         \hline
         not used & not used & .816 & .615 & .711 & .605 & .888 & .514\\
         \hline
         \multirow{3}{*}{auto-generated} & predicted from top words & .742 & .744 & .685 & .660 & .866 & .624\\
         & annotated from top words & .760 & .818 & .679 & .696 & .882 & .662\\
         
         &  annotated from whole vocabulary & .773 & \textbf{.857} & .685 &  \textbf{.726} & .752 & \textbf{.720} \\
         \hline
        
         human-generated & not used & .818 & \textbf{.869}  & .705 & \textbf{.762} & n/a & n/a \\
         \hline
    \end{tabular}
    \caption{Classification accuracy results. (CTF is human-generated counterfactual testing data.)}
    \label{tab:robust}
\end{table*}

\subsection{Causal term identification}

According to the left panel of Figure~\ref{fig:imdb_causal_terms}, we find that the similarity scores of closest opposite matches seem to be a viable signal of true causal terms. The right panel shows the performance of identifying causal terms when thresholding on the closest opposite match scores. Using threshold 0.95, we identify 32 causal terms for IMDB-L and IMDB-S datasets, of which 27 are true causal terms (i.e., precision: 84\%), and 23 causal terms for Kindle dataset, of which 19 are true causal terms (i.e., precision: 83\%).




\subsection{Robust classification for counterfactual test data}\label{sec.robust_classification}

We fit five binary LogisticRegression classifiers with different training data (using scikit-learn~\cite{scikit-learn}) and evaluate their performance on the original test samples as well as counterfactual test samples. The training data compared below have increasing requirements for human supervision. For the first and second, only original training data is required. For the third and fourth, a human provides a list of causal terms, either by selecting from the list of top terms, or from the entire vocabulary. In the final setting, humans manually annotate counterfactual training samples (equivalent to the approach of \citenoun{kaushik2019learning}). Details of the five levels of human supervision are as follows:
\begin{enumerate}
    \item Only original training samples.
    \item The original training samples are augmented with automatically generated counterfactual training samples using {\it predicted causal terms} .
    \item The original training samples are augmented with counterfactual samples automatically generated using {\it human annotated causal terms from top words} (i.e., 65 for IMDB-L, 80 for IMDB-S, and 76 for Kindle).
    \item The original training samples are augmented with counterfactual samples automatically generated using {\it human annotated causal terms from the entire vocabulary} (i.e., 282 for IMDB-L, 285 for IMDB-S, and 264 for Kindle).
    \item The original training samples are augmented with {\it human-generated counterfactual} training samples.
\end{enumerate}

We train the classifiers using the five different training sets and compare their performances on the original test samples and the human-generated counterfactual test samples. Table~\ref{tab:robust} shows the results. 

When the classifier is trained on original training samples, it performs well on the original test data, but the accuracy degrades quickly when tested on human-generated counterfactual data (e.g., 20.1\% absolute decrease for IMDB-L, 10.6\% decrease for IMDB-S, 37.4\% decrease for Kindle). This indicates that spurious correlations learned in the original classifier do not generalize well on the counterfactual test data.

When evaluating on human-generated counterfactual test samples, the classifier performance increases when we augment the original training data with counterfactual data. Even with no additional human supervision, the approach that automatically identifies causal terms outperforms the original classifier across all datasets (13\%, 5.5\%, 11\% absolute improvement). Further improvements occur with additional human supervision in the form of causal terms. Using all causal terms (less than 300 terms per dataset), the approach achieves comparable performance to the more expensive baseline which requires humans to edit $>1K$ counterfactual samples.\footnote{We lack human-generated counterfactual training samples for Kindle dataset, so we omit that result from Table~\ref{tab:robust}.}

We also observe that model accuracy slightly decreases on the original test data. This is because the spurious correlations hold in the original test data, but the importance of such features is reduced in the models trained on counterfactual samples. This suggests a potential tradeoff between accuracy on a specific dataset and generalizability of the model.


\subsection{Alternative experiments}
The Appendix\footnote{The Appendix is available in the Arxiv version of this paper.} provides additional results using more complex neural network models (LSTM with distributed word representations). The baseline classification accuracy is quite similar (within .03), and the relative accuracy of the different approaches exhibit very similar trends with the current results using logistic regression.


We have also run experiments to control for the training data size by downsampling the augmented training data to have the same size as original training data. Results show that there are only minor changes in accuracy (i.e., $<$ 0.04), and the overall trends match the current results (see Appendix).

Finally, as regularization terms may impact spurious features, we have also experimented with the L2 regularization term in logistic regression ($\{C=0.01, 0.1, 1, 10, 100\}$), and there are only minor differences in accuracy on the counterfactual test data (see Appendix).





\begin{table*}[t]
    \small
    \centering
    \begin{tabular}{|p{1.5cm}|p{1.6cm}|p{1.8cm}|p{1.7cm}|p{4.0cm}|p{3.5cm}|}
      \hline
        &\textbf{Term} & \textbf{Original coef} &\textbf{Robust coef} & \textbf{Original sentence} & \textbf{Counterfactual sentence} \\
      \hline
      \hline
       \multirow{3}{*}{\textbf{\shortstack[l]{Non-causal \\terms}}} & movie & -0.236 & 0.028 & \textbf{Terrible} movie & \textbf{Fantastic} movie \\
      \cline{2-6}
       & free & -1.41 & -0.919 & This was a free book that sounded \textbf{boring} to me. & This was a free book that sounded \textbf{interesting} to me. \\
       \hline
       \hline
       
       \multirow{3}{*}{\textbf{\shortstack[l]{Causal \\terms}}} & awesome & 0.584 & 1.838 & He was an \textbf{awesome} actor. & He was an \textbf{awful} actor. \\
       \cline{2-6}
       & terrible & -1.283 & -2.336 & The whole movie consists of \textbf{terrible} dialogue. & The whole movie consists of \textbf{pleasant} dialogue. \\
       \cline{2-6}
       \hline
    \end{tabular}
    \caption{Coefficient change of causal and non-causal terms. }
    \label{tab.spurious_coef}
\end{table*}

\subsection{Performance change with different number of human annotated causal terms}

\begin{figure}[t]
	\centering
	\includegraphics[width=3.5in]{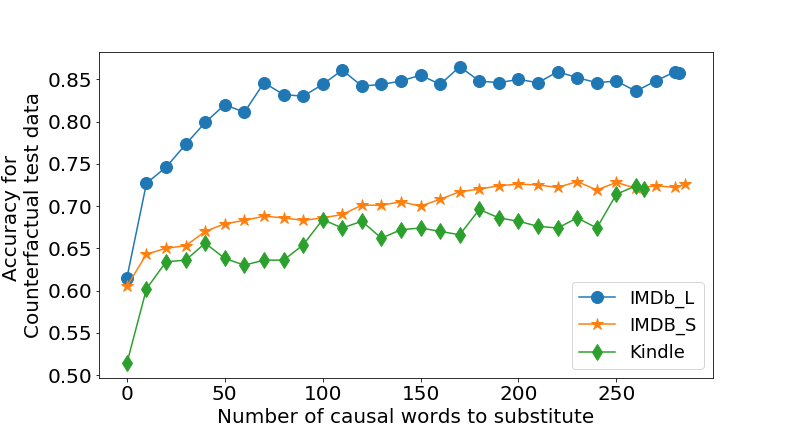}
	\caption{Performance change with counterfactuals generated using different numbers of annotated causal terms}
	\label{fig:trend}
\end{figure}
To further investigate how many human-provided causal terms are needed to improve robustness, Figure~\ref{fig:trend} shows the classification performance with different numbers of causal terms used for generating counterfactual samples. The quality of automatically generated counterfactuals depends on the causal terms used for antonym substitutions. We observe that performance seems to plateau after about 100 causal terms, which suggests that we can get similar performance by annotating 100 causal terms, as opposed to creating $>1K$ counterfactual training samples. The cause of the plateau is likely due to the infrequency of subsequent terms and the fact that such terms co-occur with other causal terms, so they do not result in many new counterfactual samples.

\begin{table}[t]
    \small
    \centering
    \begin{tabular}{|p{3.2cm}|p{2.1cm}|p{2.0cm}|}
      \hline
        \textbf{Corrected samples} & \textbf{Original coef} &\textbf{Robust coef}\\
      \hline
       Really good movie.(pos) & \makecell[l]{good:0.231 \\movie:-0.236} & \makecell[l]{good:0.714 \\movie:0.028} \\
       \hline
       The dubbing was as good as I have seen.(pos) & \makecell[l]{dubbing:-0.472 \\good:0.231} & \makecell[l]{dubbing:-0.1 \\ good:0.714} \\
       \hline
       The story was incredibly interesting.(pos) & \makecell[l]{story:-0.171 \\  incredibly:-0.874 \\ interesting:-0.874} & \makecell[l]{story:-0.083 \\ incredibly:0.029 \\ interesting:1.012}\\
       \hline
    \end{tabular}
    \caption{Explanation for robust classifier corrected samples.}
    \label{tab.coef_change}
\end{table}


\begin{table}[t]
    \small
    \centering
    \begin{tabular}{|c|c|c|c|c|}
      \hline
 &\multicolumn{2}{c|}{\textbf{Change per document}}& \multicolumn{2}{c|}{\textbf{ Change per term}}\\
        \cline {2-5}
        & \textbf{causal}& \textbf{non-causal} & \textbf{causal}&\textbf{non-causal}\\
      \hline
       \textbf{IMDB-L} & 1.888 & -0.734 & 0.327 & -0.01 \\
      \hline
       \textbf{IMDB-S} & 0.435 & -0.626 & 0.302& -0.042 \\
      \hline
       \textbf{Kindle} & 0.293 & -0.772 & 0.315 & -0.109 \\
       \hline
    \end{tabular}
    \caption{Original versus robust classifier coefficient change for causal versus non-causal terms for corrected samples.}
    \label{tab.impact}
\end{table}

\subsection{Coefficient change for causal and non-causal terms}

To understand why training with counterfactual data improves classifier robustness, Table~\ref{tab.spurious_coef} shows examples of the change in coefficients from the original classifier to the robust classifier for causal and non-causal terms.

Take the word ``free'' for example, which has a coefficient $-1.41$ in the original classifier due to its high frequency in negative samples. When generating counterfactuals, we substitute ``boring'' with ``interesting'' and generate a sample ``This was a free book that sounded \textit{interesting} to me'' with positive sentiment. So the counterfactual samples increase the frequency of ``free'' in positive class and thus mitigate the spurious correlation of ``free'' with negative class. The classifier trained on a combination of original data and counterfactual data would learn a smaller magnitude coefficient for ``free''. Analogously, the approach increases the magnitude of causal terms such as ``awesome'' and ``awful'' by providing counterfactuals with opposite labels.

\subsection{Error analysis}
Table~\ref{tab.coef_change} shows several test sentences that are misclassified by the original classifier and later corrected by the robust classifier. We can see again that the robust classifier increases coefficients of causal terms and decreases coefficients of non-causal terms. For example, ``Really good movie'' is incorrectly predicted as negative by the original classifier, because the causal term ``good'' has a small positive coefficient and the prediction is misled by the spuriously correlated negative term ``movie''\footnote{In the data, ``film'' correlates with high ratings, while ``movie'' correlates with low ratings.}. The robust classifier corrects this prediction by increasing the coefficient of the causal term ``good'' and decreasing the coefficient of the non-causal term ``movie.''

We conduct a final analysis to explore the impact of causal versus non-causal terms when correcting misclassifications. For each corrected sample, we compute separately the change in coefficient magnitudes for causal and non-causal terms. We then aggregate across all corrected samples to summarize the impact each type of correction has.  As shown in Table~\ref{tab.impact}, for IMDB-L, increasing coefficients of causal terms is more important than decreasing coefficients of non-causal terms, and the reverse is true for the other two datasets. This suggests that document length is an important factor in determining whether increasing coefficients of causal terms has bigger impacts or decreasing coefficients of non-causal terms has bigger impacts. Examining the average coefficient change of each term, the robust classifier tends to make bigger increases for causal terms and smaller decreases for non-causal terms. However, the greater frequency of non-causal terms can lead to these changes to have a greater overall impact on classification accuracy.

\section{Conclusion and Future Work}
We have presented a framework to automatically generate counterfactual training samples from causal terms and then train a robust classifier using the combination of original data and counterfactual data. Using this framework, we can easily improve classifier robustness even with few causal terms. If enough causal terms are annotated (e.g., 100 in our experiments), it is possible to achieve performance comparable to using human-generated counterfactuals. In future work, we will investigate extensions to increase the precision and recall of causal term identification to further reduce the reliance on human supervision. Additionally, it would be interesting to extend this framework to other tasks such as topic classification robustness. To do so, we would need to generalize the notion of ``antonyms'' to include terms that indicate a different topic (e.g., to convert a sports news story to a political news story, we might change the sentence ``watch the game'' to ``watch the debate''). Then we could generate ``counterfactuals'' by substituting topic-related terms with terms that are not semantically related to the current topic (or related to other topics).



\section*{Acknowledgments}
This  research  was  funded  in  part  by  the  National  Science Foundation under grant \#1618244. Zhao Wang was funded in part by a Dissertation Fellowship from the Computer Science department at Illinois Tech. We would also like to thank the anonymous reviewers for useful feedback.

\bibliography{reference}

\begin{thebibliography}{32}
\providecommand{\natexlab}[1]{#1}
\providecommand{\url}[1]{\texttt{#1}}
\providecommand{\urlprefix}{URL }
\expandafter\ifx\csname urlstyle\endcsname\relax
  \providecommand{\doi}[1]{doi:\discretionary{}{}{}#1}\else
  \providecommand{\doi}{doi:\discretionary{}{}{}\begingroup
  \urlstyle{rm}\Url}\fi

\bibitem[{Dai, Chen, and Li(2019)}]{Dai2019ABA}
Dai, J.; Chen, C.; and Li, Y. 2019.
\newblock A Backdoor Attack Against LSTM-Based Text Classification Systems.
\newblock \emph{IEEE Access} 7: 138872--138878.

\bibitem[{Devlin et~al.(2019)Devlin, Chang, Lee, and
  Toutanova}]{Devlin2019BERTPO}
Devlin, J.; Chang, M.-W.; Lee, K.; and Toutanova, K. 2019.
\newblock BERT: Pre-training of Deep Bidirectional Transformers for Language
  Understanding.
\newblock In \emph{NAACL-HLT}.

\bibitem[{Forman(2003)}]{Forman2003}
Forman, G. 2003.
\newblock An Extensive Empirical Study of Feature Selection Metrics for Text
  Classification.
\newblock \emph{J. Mach. Learn. Res.} 3: 1289–1305.
\newblock ISSN 1532-4435.

\bibitem[{Garg et~al.(2019)Garg, Perot, Limtiaco, Taly, Chi, and
  Beutel}]{garg2019counterfactual}
Garg, S.; Perot, V.; Limtiaco, N.; Taly, A.; Chi, E.~H.; and Beutel, A. 2019.
\newblock Counterfactual Fairness in Text Classification through Robustness.
\newblock In \emph{Proceedings of the 2019 AAAI/ACM Conference on AI, Ethics,
  and Society}, AIES'19.

\bibitem[{Ghorbani, Abid, and Zou(2019)}]{ghorbani2019interpretation}
Ghorbani, A.; Abid, A.; and Zou, J. 2019.
\newblock Interpretation of neural networks is fragile.
\newblock In \emph{Proceedings of the AAAI Conference on Artificial
  Intelligence}, volume~33, 3681--3688.

\bibitem[{He and McAuley(2016)}]{He_2016}
He, R.; and McAuley, J. 2016.
\newblock Ups and Downs.
\newblock \emph{Proceedings of the 25th International Conference on World Wide
  Web - WWW'16} \doi{10.1145/2872427.2883037}.

\bibitem[{Hoerl and Kennard(1970)}]{Hoerl1970Regression}
Hoerl, A.~E.; and Kennard, R.~W. 1970.
\newblock Ridge Regression: Biased Estimation for Nonorthogonal Problems.
\newblock \emph{Technometrics} 12(1): 55--67.
\newblock \doi{10.1080/00401706.1970.10488634}.

\bibitem[{Imbens(2004)}]{imbens2004nonparametric}
Imbens, G.~W. 2004.
\newblock Nonparametric estimation of average treatment effects under
  exogeneity: A review.
\newblock \emph{Review of Economics and statistics} 86(1): 4--29.

\bibitem[{Jia et~al.(2019)Jia, Raghunathan, G{\"o}ksel, and
  Liang}]{jia2019certified}
Jia, R.; Raghunathan, A.; G{\"o}ksel, K.; and Liang, P. 2019.
\newblock Certified Robustness to Adversarial Word Substitutions.
\newblock In \emph{Proceedings of the 2019 Conference on Empirical Methods in
  Natural Language Processing and the 9th International Joint Conference on
  Natural Language Processing (EMNLP-IJCNLP)}, 4129--4142. Hong Kong, China.

\bibitem[{Kaufman, Rosset, and Perlich(2011)}]{Shachar2011Leakage}
Kaufman, S.; Rosset, S.; and Perlich, C. 2011.
\newblock Leakage in Data Mining: Formulation, Detection, and Avoidance.
\newblock In \emph{Proceedings of the 17th ACM SIGKDD International Conference
  on Knowledge Discovery and Data Mining}, KDD '11, 556–563. New York, NY,
  USA.
\newblock ISBN 9781450308137.

\bibitem[{Kaushik, Hovy, and Lipton(2020)}]{kaushik2019learning}
Kaushik, D.; Hovy, E.; and Lipton, Z. 2020.
\newblock Learning The Difference That Makes A Difference With
  Counterfactually-Augmented Data.
\newblock In \emph{International Conference on Learning Representations},
  ICLR'20.

\bibitem[{Keith, Jensen, and O{'}Connor(2020)}]{keith2020text}
Keith, K.; Jensen, D.; and O{'}Connor, B. 2020.
\newblock Text and Causal Inference: A Review of Using Text to Remove
  Confounding from Causal Estimates.
\newblock In \emph{Proceedings of the 58th Annual Meeting of the Association
  for Computational Linguistics}, 5332--5344. Online: ACL.

\bibitem[{King and Nielsen(2019)}]{king2019propensity}
King, G.; and Nielsen, R. 2019.
\newblock Why propensity scores should not be used for matching.
\newblock \emph{Political Analysis} 27(4): 435--454.

\bibitem[{Kiritchenko and Mohammad(2018)}]{kiritchenko2018examining}
Kiritchenko, S.; and Mohammad, S. 2018.
\newblock Examining Gender and Race Bias in Two Hundred Sentiment Analysis
  Systems.
\newblock In \emph{Proceedings of the Seventh Joint Conference on Lexical and
  Computational Semantics}, 43--53. New Orleans, Louisiana: ACL.
\newblock \doi{10.18653/v1/S18-2005}.

\bibitem[{Landeiro and Culotta(2018)}]{Virgile2018confounding}
Landeiro, V.; and Culotta, A. 2018.
\newblock Robust Text Classification under Confounding Shift.
\newblock \emph{Journal of Artificial Intelligence Research} 63: 391--419.
\newblock \doi{10.1613/jair.1.11248}.

\bibitem[{Lu et~al.(2018)Lu, Mardziel, Wu, Amancharla, and
  Datta}]{lu2018gender}
Lu, K.; Mardziel, P.; Wu, F.; Amancharla, P.; and Datta, A. 2018.
\newblock Gender Bias in Neural Natural Language Processing.
\newblock In \emph{Logic, Language, and Security}, volume 12300, 189--202.
  Springer, Cham.
\newblock \doi{https://doi.org/10.1007/978-3-030-62077-6_14}.

\bibitem[{Martens and Provost(2014)}]{martens2014explaining}
Martens, D.; and Provost, F. 2014.
\newblock Explaining data-driven document classifications.
\newblock \emph{Mis Quarterly} 38(1): 73--100.

\bibitem[{Pang and Lee(2005)}]{pang2005seeing}
Pang, B.; and Lee, L. 2005.
\newblock Seeing stars: Exploiting class relationships for sentiment
  categorization with respect to rating scales.
\newblock In \emph{Proceedings of the 43rd annual meeting on association for
  computational linguistics}, 115--124. ACL.

\bibitem[{Paul(2017)}]{paul-2017-feature}
Paul, M.~J. 2017.
\newblock Feature Selection as Causal Inference: Experiments with Text
  Classification.
\newblock In \emph{Proceedings of the 21st Conference on Computational Natural
  Language Learning ({C}o{NLL} 2017)}. Vancouver, Canada: ACL.

\bibitem[{Pedregosa et~al.(2011)Pedregosa, Varoquaux, Gramfort, Michel,
  Thirion, Grisel, Blondel, Prettenhofer, Weiss, Dubourg, Vanderplas, Passos,
  Cournapeau, Brucher, Perrot, and Duchesnay}]{scikit-learn}
Pedregosa, F.; Varoquaux, G.; Gramfort, A.; Michel, V.; Thirion, B.; Grisel,
  O.; Blondel, M.; Prettenhofer, P.; Weiss, R.; Dubourg, V.; Vanderplas, J.;
  Passos, A.; Cournapeau, D.; Brucher, M.; Perrot, M.; and Duchesnay, E. 2011.
\newblock Scikit-learn: Machine Learning in {P}ython.
\newblock \emph{Journal of Machine Learning Research} 12: 2825--2830.

\bibitem[{Quionero-Candela et~al.(2009)Quionero-Candela, Sugiyama,
  Schwaighofer, and Lawrence}]{Joaquin2009datasetshift}
Quionero-Candela, J.; Sugiyama, M.; Schwaighofer, A.; and Lawrence, N.~D. 2009.
\newblock \emph{Dataset Shift in Machine Learning}.
\newblock The MIT Press.
\newblock ISBN 0262170051.

\bibitem[{Ribeiro et~al.(2020)Ribeiro, Wu, Guestrin, and
  Singh}]{ribeiro2020accuracy}
Ribeiro, M.~T.; Wu, T.; Guestrin, C.; and Singh, S. 2020.
\newblock Beyond Accuracy: Behavioral Testing of {NLP} Models with
  {C}heck{L}ist.
\newblock In \emph{Proceedings of the 58th Annual Meeting of the Association
  for Computational Linguistics}. ACL.

\bibitem[{Roemmele, Bejan, and Gordon(2011)}]{Melissa2011causal}
Roemmele, M.; Bejan, C.; and Gordon, A. 2011.
\newblock Choice of Plausible Alternatives: An Evaluation of Commonsense Causal
  Reasoning.
\newblock In \emph{AAAI Spring Symp - Technical Report}.

\bibitem[{Sagawa et~al.(2020)Sagawa, Raghunathan, Koh, and
  Liang}]{sagawa2020investigation}
Sagawa, S.; Raghunathan, A.; Koh, P.~W.; and Liang, P. 2020.
\newblock An Investigation of Why Overparameterization Exacerbates Spurious
  Correlations.
\newblock In \emph{Proceedings of the 37th International Conference on Machine
  Learning,{ICML} 2020}.

\bibitem[{Sap et~al.(2018)Sap, Bras, Allaway, Bhagavatula, Lourie, Rashkin,
  Roof, Smith, and Choi}]{sap2018atomic}
Sap, M.; Bras, R.~L.; Allaway, E.; Bhagavatula, C.; Lourie, N.; Rashkin, H.;
  Roof, B.; Smith, N.~A.; and Choi, Y. 2018.
\newblock ATOMIC: An Atlas of Machine Commonsense for If-Then Reasoning.
\newblock \emph{ArXiv} abs/1811.00146.

\bibitem[{Srivastava, Hashimoto, and Liang(2020)}]{srivastava2020robustness}
Srivastava, M.; Hashimoto, T.; and Liang, P. 2020.
\newblock Robustness to Spurious Correlations via Human Annotations.
\newblock In III, H.~D.; and Singh, A., eds., \emph{Proceedings of the 37th
  International Conference on Machine Learning}, volume 119 of
  \emph{Proceedings of Machine Learning Research}, 9109--9119.

\bibitem[{Wang and Culotta(2020)}]{wang-culotta-2020-identifying}
Wang, Z.; and Culotta, A. 2020.
\newblock Identifying Spurious Correlations for Robust Text Classification.
\newblock In \emph{Findings of the Association for Computational Linguistics,
  EMNLP 2020}.

\bibitem[{Winship and Morgan(1999)}]{winship1999estimation}
Winship, C.; and Morgan, S.~L. 1999.
\newblock The estimation of causal effects from observational data.
\newblock \emph{Annual review of sociology} 25(1): 659--706.

\bibitem[{Wood-Doughty, Shpitser, and Dredze(2018)}]{wooddoughty2018challenges}
Wood-Doughty, Z.; Shpitser, I.; and Dredze, M. 2018.
\newblock Challenges of Using Text Classifiers for Causal Inference.
\newblock \emph{Proceedings of the Conference on Empirical Methods in Natural
  Language Processing. EMNLP} 2018: 4586--4598.

\bibitem[{Wulczyn, Thain, and Dixon(2017)}]{Ellery2017toxic}
Wulczyn, E.; Thain, N.; and Dixon, L. 2017.
\newblock Ex Machina: Personal Attacks Seen at Scale.
\newblock In \emph{Proceedings of the 26th International Conference on World
  Wide Web}, WWW '17.

\bibitem[{Zaidan and Eisner(2008)}]{Zaidan2008ModelingAA}
Zaidan, O.~F.; and Eisner, J. 2008.
\newblock Modeling Annotators: A Generative Approach to Learning from Annotator
  Rationales.
\newblock In \emph{Proceedings of the Conference on Empirical Methods in
  Natural Language Processing}, EMNLP '08. ACL.

\bibitem[{Zmigrod et~al.(2019)Zmigrod, Mielke, Wallach, and
  Cotterell}]{zmigrod-etal-2019-counterfactual}
Zmigrod, R.; Mielke, S.~J.; Wallach, H.; and Cotterell, R. 2019.
\newblock Counterfactual Data Augmentation for Mitigating Gender Stereotypes in
  Languages with Rich Morphology.
\newblock In \emph{Proceedings of the 57th Annual Meeting of the Association
  for Computational Linguistics}. ACL.

\end{thebibliography}

\clearpage
\appendix
\onecolumn
\section{Appendix}
\label{sec:appendix}

We provide supplemental information in this section.

\subsection{Results using neural network models}
Besides the LogisticRegression classifier, we also experiment with the LSTM binary classifier that takes an embedding layer initialized from the pre-trained word2vec model as input, using the Adam optimizer, binary cross-entropy loss, a batch size of 32, and 10 epochs. Table~\ref{tab:lstm} shows the results. Comparing results from Table~\ref{tab:lstm} and Table~\ref{tab:robust}: (i) the baseline accuracies for the three testing datasets are quite similar (within 0.05). One exception is that the Kindle counterfactual test set gains the most improvement (about 0.12), which might be due to the quality and amount of automatically generated counterfactuals; (ii) the relative accuracies of the five different approaches with LSTM exhibit very similar trends with results using LogisticRegression: augmenting with counterfactual data produced by increasing levels of human supervision help improve the model performance. While more sophisticated representations can improve overall results, we wanted to first understand model behavior in a simple and interpretable setting and avoid possible overfitting with complex models.

\begin{table*}[h]
    \centering
    \begin{tabular}{|p{2.8cm}|p{5.2cm}||p{.7cm}|p{.7cm}||p{.7cm}|p{.7cm}||p{.7cm}|p{.7cm}|} 
    \hline
     \multicolumn{2}{|c||}{\textit{Training data:}} & 
         \multicolumn{6}{c|}{\textit{Testing data}} \\
         \cline{3-8}
         \multicolumn{2}{|c||}{\textit{Original train samples + Counterfactual train samples}} &
         \multicolumn{2}{c||}{\textbf{IMDB-L}} & \multicolumn{2}{c||}{\textbf{IMDB-S}} & \multicolumn{2}{c|}{\textbf{Kindle}}\\
      
         \hline
         {\bf Counterfactual training samples} & {\bf Causal terms} & {\bf Orig} & {\bf CTF} & {\bf Orig} & {\bf CTF} & {\bf Orig} & {\bf CTF} \\
         \hline
         not used & not used & .811 & .700 & .741 & .656 & .908 & .642\\
         \hline
         \multirow{3}{*}{auto-generated} & predicted from top words & .691 & .737 & .728 & .702 & .896 & .712\\
         & annotated from top words & .721 & .752 & .715 & .725 & .894 & .760\\
         
         &  annotated from whole vocabulary & .754 & \textbf{.803} & .717 &  \textbf{.751} & .864 & \textbf{.840} \\
         \hline
        
         human-generated & not used & .814 & \textbf{.867}  & .743 & \textbf{.785} & n/a & n/a \\
         \hline
    \end{tabular}
    \caption{Classification accuracy results using LSTM. (CTF is human-generated counterfactual testing data.)}
    \label{tab:lstm}
\end{table*}

\subsection{Regularization}
To explore the impact of regularization on model robustness, we experiment with L2 regularization (strength controlled by C) in LogisticRegression. Table~\ref{tab:regularization} shows the performance of LogisticRegression classifier fitted on original training samples with the regularization strength varying by C=0.01, 0.1, 1, 10, 100. There's only minor improvements (within 0.02) with a careful choice of regularization. Since many spurious features have very high correlations with the classes, regularization will not be able to reduce their impact sufficiently.

\begin{table*}[h]
    \centering
    \begin{tabular}{|p{3.5cm}||p{.7cm}|p{.7cm}||p{.7cm}|p{.7cm}||p{.7cm}|p{.7cm}|} 
    \hline
     {\textit{Training data:}} & 
         \multicolumn{6}{c|}{\textit{Testing data}} \\
         \cline{2-7}
         \textit{Original train samples } &
         \multicolumn{2}{c||}{\textbf{IMDB-L}} & \multicolumn{2}{c||}{\textbf{IMDB-S}} & \multicolumn{2}{c|}{\textbf{Kindle}}\\
      
         \hline
         {\bf C} & {\bf Orig} & {\bf CTF} & {\bf Orig} & {\bf CTF} & {\bf Orig} & {\bf CTF} \\
         \hline
         0.01 & .795 & .578 & .689 & .554 & .826 & .408 \\
         
         0.1 & .811 & .605 & {\bf .722} & .595 & .884 & .452 \\
         1 & {\bf .816} & .615 & .711 & {\bf .605} & {\bf .888} & .514 \\
         10 & .809 & {\bf .617} & .668 & .598 & .856 & {\bf .534} \\
         100 & .797 & .611 & .643 & .573 & .846 & .520 \\
         \hline
    \end{tabular}
    \caption{LogisticRegression classifier accuracy with L2 regularization controlled by C.}
    \label{tab:regularization}
\end{table*}

\subsection{Control the size of training data}
In section $6.2$, we conduct experiments to fit binary classifiers using training data that have five levels of increasing requirements for human supervision. The size of the training data is changing when the original training data is augmented with counterfactual data. To isolate the impact of training data size has on model performance, we run experiments to control for training data size by downsampling the augmented training data to have the same size as the original training data. Table~\ref{tab:control_size} shows the classification accuracy with controlled training data size. Comparing with results from Table~\ref{tab:robust}, there are only minor changes in accuracy (within 0.04), and the overall trends match the other results (i.e., training with counterfactual data improves classifier robustness).

\begin{table*}[h]
    \centering
    \begin{tabular}{|p{2.8cm}|p{5.2cm}||p{.7cm}|p{.7cm}||p{.7cm}|p{.7cm}||p{.7cm}|p{.7cm}|} 
    \hline
     \multicolumn{2}{|c||}{\textit{Training data:}} & 
         \multicolumn{6}{c|}{\textit{Testing data}} \\
         \cline{3-8}
         \multicolumn{2}{|c||}{\textit{Original train samples + Counterfactual train samples}} &
         \multicolumn{2}{c||}{\textbf{IMDB-L}} & \multicolumn{2}{c||}{\textbf{IMDB-S}} & \multicolumn{2}{c|}{\textbf{Kindle}}\\
      
         \hline
         {\bf Counterfactual training samples} & {\bf Causal terms} & {\bf Orig} & {\bf CTF} & {\bf Orig} & {\bf CTF} & {\bf Orig} & {\bf CTF} \\
         \hline
         not used & not used & .816 & .615 & .711 & .605 & .888 & .514\\
         \hline
         \multirow{3}{*}{auto-generated} & predicted from top words & .648 & .713 & .670 & .659 & .816 & .614\\
         & annotated from top words & .689 & {\bf .803} & .644 & {\bf .696} & .812 & .626\\
         
         &  annotated from whole vocabulary & .701 & .787 & .649 &  .685 & .740 & \textbf{.706} \\
         \hline
        
         human-generated & not used & .807 & \textbf{.848}  & .697 & \textbf{.732} & n/a & n/a \\
         \hline
    \end{tabular}
    \caption{Classification accuracy with controlled training data size. Training with counterfactual data improves performance.}
    \label{tab:control_size}
\end{table*}

\end{document}